\documentclass[11pt, a4paper, logo, onecolumn, internal, copyright,]{googledeepmind}

\usepackage[authoryear, sort&compress, round]{natbib}
\usepackage[T1]{fontenc}
\usepackage[utf8]{inputenc}
\usepackage{todonotes}
\usepackage{algorithm}
\usepackage{algpseudocode}
\usepackage{graphicx}
\usepackage{wrapfig}
\usepackage{multirow}
\title{Learning to Learn from Language Feedback with Social Meta-Learning}

\renewcommand{\today}{2025-09-24}

\author[1]{Jonathan Cook}
\author[1]{Diego Antognini}
\author[1]{Martin Klissarov}
\author[1]{Claudiu Musat}
\author[1]{Edward Grefenstette}

\affil[1]{Google DeepMind}

\usepackage{listings}      
\usepackage[most,skins,theorems]{tcolorbox}

\definecolor{darkpurple}{RGB}{102, 51, 153}
\definecolor{lightpurple}{RGB}{204, 153, 255}
\definecolor{lightblue}{rgb}{0.22,0.45,0.70}
\definecolor{forestgreen}{rgb}{0.24,0.50,0.19}

\definecolor{DeepTeal}{RGB}{20, 110, 110}  
\definecolor{LightTeal}{RGB}{235, 245, 245} 

\newcolumntype{L}[1]{>{\raggedright\arraybackslash}p{#1}}

\tcbset{
  aibox/.style={
    width=\linewidth,
    top=8pt,
    bottom=4pt,
    colback=LightTeal,
    colframe=DeepTeal,
    colbacktitle=DeepTeal,
    fonttitle=\bfseries,
    enhanced,
    center,
    attach boxed title to top left={yshift=-0.1in,xshift=0.15in},
    boxed title style={boxrule=0pt,colframe=white,},
  }
}
\newtcolorbox{AIbox}[2][]{aibox,title=#2,#1}

\begin{abstract}
Large language models (LLMs) often struggle to learn from corrective feedback within a conversational context. They are rarely proactive in soliciting this feedback, even when faced with ambiguity, which can make their dialogues feel static, one-sided, and lacking the adaptive qualities of human conversation. To address these limitations, we draw inspiration from social meta-learning (SML) in humans --- the process of learning how to learn from others. We formulate SML as a finetuning methodology, training LLMs to solicit and learn from language feedback in simulated pedagogical dialogues, where static tasks are converted into interactive social learning problems. SML effectively teaches models to use conversation to solve problems they are unable to solve in a single turn. This capability generalises across domains; SML on math problems produces models that better use feedback to solve coding problems and vice versa. Furthermore, despite being trained only on fully-specified problems, these models are better able to solve underspecified tasks where critical information is revealed over multiple turns. When faced with this ambiguity, SML-trained models make fewer premature answer attempts and are more likely to ask for the information they need. This work presents a scalable approach to developing AI systems that effectively learn from language feedback. 
\end{abstract}

\begin{document}
\maketitle
\section{Introduction}
The broad capabilities of large language models (LLMs) have led to their widespread integration into countless applications. Often, these applications are conversational, positioning the models as dialogue partners. However, the interactive nature of these settings is not actively used by current models. Users of chat-based LLMs report experiencing restrictive dialogues and a low degree of adaptive freedom as conversations progress \citep{Wang2024Understanding}. Moreover, models are rarely proactive in soliciting feedback when faced with ambiguity \citep{Li2025Quest, Zhang2025Modeling}. 

In children, the challenge of learning via social interaction is itself thought to be learned through a process referred to as social meta-learning (SML) \citep{Allen2017Social}. By adulthood, humans are generally adept social learners, using others as a learning resource and effectively engaging in collaborative problem solving. Meanwhile, current approaches to LLM post-training tend to focus on single-turn reasoning and static task performance, limiting opportunities for learning to learn from contextual feedback and possibly diminishing the conversational adaptation abilities already present within the pre-trained base model \citep{Shaikh2023Grouding, Wang2024MINT}. Interactions with LLMs can subsequently have a brittleness that is unfamiliar within interactions with other humans, placing a significant burden of initial prompt engineering on everyday users \citep{White2023A}.

To address this problem, we formulate SML as a finetuning methodology for LLMs. We achieve this by converting static tasks, such as math problems, into interactive, pedagogical dialogues. Given an initial problem statement, a ``student'' model attempts to generate the solution over the course of a conversation and a ``teacher'' model provides guidance. The student is the model being trained and the teacher can be a frozen instance of the same model, or a stronger model. Crucially, the teacher is provided with privileged information, such as the correct final answer or access to the outputs of a verifier. This creates an information asymmetry, ensuring that the teacher can provide valuable, corrective feedback and making problems that are significantly beyond the student's single-turn capabilities tractable through interaction. It also incentivises the student to be proactive in extracting relevant information from the teacher, analogous to in-context exploration in partially observable sequential decision making problems \citep{Dai2024In-context, Krishnamurthy2024Can}.

We explore two approaches to SML: an offline method where we gather a filtered dataset of successful dialogues and perform supervised finetuning (SFT), and online RL using binary, conversation-level rewards. We find that online RL yields much greater improvements in the ability to learn from language feedback at test time. Moreover, this ability generalises to longer conversations at test time than were used for training and transfers across domains; SML on math problems leads to improved performance in learning from language feedback on coding tasks. 

Crucially for human-AI interaction, this training paradigm also enhances a model's ability to navigate ambiguity. Despite the fact that our SML setup exclusively involves training on problems that are fully specified from the first turn, the finetuned model achieves superior performance on tasks where critical information is revealed over multiple conversational turns. A desirable property of models in this setting is making fewer premature answer attempts and instead asking for the necessary information. While this behaviour does become more frequent from SML alone, we find that it can be significantly enhanced through a two-stage training process. We introduce Q-priming, a preliminary SFT stage where we train the model on dialogues in which we have explicitly prompted it to ask questions. To generate examples of useful questions, we again leverage the information asymmetry of our setup, providing the model with the teacher's private knowledge (e.g., the ground truth solution) and asking it to formulate an informative query. Q-priming encourages exploratory, information-seeking behaviour, which is then refined by subsequent online RL training.

In summary, we introduce a novel finetuning methodology directed at enhancing the adaptability of LLMs by improving their use of language feedback. This explicitly frames conversation as an in-context learning environment. Our contributions are as follows:

\begin{enumerate}
    \item We formulate SML as a finetuning methodology  (Section \ref{sec:SML}) and demonstrate that it is more effective than single-turn finetuning at teaching models to learn from language feedback (Section \ref{sec:results_1_2}). In comparing training strategies, we find that online RL is substantially more effective than performing offline data generation and SFT.
    \item We show that the ability to learn from language feedback, acquired via SML, generalises across domains. Specifically, a model trained with SML on math problems demonstrates and improved ability to learn from language feedback on coding tasks (Section \ref{sec:results_3}).
    \item We find that SML enhances a model's ability to handle ambiguity and generalise to different feedback styles. Models trained on fully-specified problems with corrective feedback show improved performance on underspecified tasks where information is revealed incrementally (Section \ref{sec:results_4}). Paired with Q-priming, this training fosters strategic exploration, leading models to ask for necessary clarifications and proactively solicit feedback (Section \ref{seq:behaviour}).
\end{enumerate}

By reframing static tasks as interactive learning opportunities, this work presents a scalable path towards developing LLMs that are not just powerful instruction followers, but collaborative assistants. This shift is essential for building more robust, helpful, and human-compatible AI systems.

\section{Preliminaries}
\label{prelims}

\paragraph{Offline Reinforcement Learning.} In this work, we make use of a finetuning strategy that involves generating a static dataset which is filtered according to a reward function and used for supervised finetuning (SFT) \citep{Zelikman2022STaR, Dong2023RAFT}. This can be considered a form of offline reinforcement learning (RL). For a given prompt $x_i \in \mathcal{D}_\text{prompt}$, the model being trained generates one or more candidate solutions, $\{y_{i,k}\} \sim \mathcal{M}_S(x_i)$. These generated solutions are then evaluated by a reward function or external verifier $R(x, y)$. A training dataset is then constructed by filtering these self-generated solutions, as exemplified for the case of binary rewards in Equation \ref{eq:star}.

\begin{equation}
    \mathcal{D}_\text{train} = \{(x_i, y_{i,k}) | R(x_i, y_{i,k}) = 1\}
    \label{eq:star}
\end{equation}

This cycle of generation, filtering, and finetuning can be iterated to progressively enhance the model's performance. In this work, we only ever use a single iteration.

\paragraph{Online Reinforcement Learning.} Instead of operating in distinct cycles of data generation and training, online RL updates the model's parameters directly in response to rewards. In this setting, the model's policy $\pi_{\theta_S}$ maximises the expected reward from a reward function $R(x, y)$. 

GRPO \citep{Shao2024DeepSeekMath} is a common online RL finetuning algorithm that uses a Monte-Carlo value estimate, alleviating the need for a critic model. $\pi_{\theta_S}$ generates a ``group'' of $G > 1$ solution candidates $\{y_{i,k}\}^G_{k=1}$ for each sampled prompt $x_i$ and is then optimised via Equation \ref{eq:grpo}.

\begin{equation}
    J(\theta_S) = \mathbb{E}_{x \sim \mathcal{D}_\text{prompt}, \{ y \}^G_1 \sim \pi_{\theta_S}(\cdot | x)} \left[ \frac{1}{G} \sum^G_{k=1} (A_k - \beta \, \mathbb{D}_\text{KL} (\pi_{\theta_S} || \pi_\text{ref})) \right]
    \label{eq:grpo}
\end{equation}

Here, $\beta$ is a hyperparameter that controls the strength of the KL penalty with respect to a reference model. The advantages $A_k$ correspond to the group normalised reward $r_k$ of each solution $y_k$, as shown in Equation \ref{eq:grpo_adv}.

\begin{equation}
    A_k = \frac{r_k - \text{mean}(\{r\}^G_1)}{\text{std}(\{r\}^G_1)}
    \label{eq:grpo_adv}
\end{equation}

\section{Related Work}
\label{related_work}

Our work on Social Meta-Learning (SML) synthesises concepts from three areas of research: training LLMs for multi-turn interaction, learning from natural language feedback, and social learning via RL. SML uniquely integrates these threads by training LLMs that learn \textit{how} to learn from social interaction, thereby becoming more effective conversational partners.

\subsection{Multi-Turn Interaction with LLMs}
\label{mt_interaction}

The shift from single-turn queries to multi-turn interactions represents a critical step towards more capable and collaborative LLMs \citep{Yi2024Survey, Li2025Beyond}. However, maintaining coherence and effectiveness over dialogues remains a significant challenge. Prior work shows that models ``get lost'' in conversational contexts \citep{Laban2025LLMs} and that common approaches to post-training might \textit{reduce} their ability to make use of language feedback \citep{Shaikh2023Grouding, Wang2024MINT}. 

A growing body of work deliberately designs finetuning algorithms aimed at improving multi-turn interaction abilities. A number of these propose novel multi-turn reinforcement learning (RL) algorithms \citep{Zhou2024ArCHer, Zhou2025SWEET} aimed at addressing core problems such as credit assignment through the use of turn-level critics. \citet{Wu2025CollabLLM} train models to collaboratively interact with users by searching over unrolled conversations with a simulated user and applying ``multiturn-aware'' rewards that combine token efficiency metrics with LLM-based judgements of interactivity. Relatedly, some work focuses specifically on training and evaluating for question asking behaviour in LLMs \citep{Andukuri2024STaR, Li2025Quest, Zhang2025Modeling}. Clarification seeking could be seen as a multi-turn analogue to ``cognitive behaviours'' such as backtracking that have proven useful for improving single-turn reasoning abilities \citep{Ghandi2025Cognitive}. Meanwhile, \citet{Chen2025Broaden} pose conversation as a planning problem and perform conversation planning in semantic space.

A parallel line of work has focused on improving non-conversational reasoning abilities through multi-turn self-correction \citep{Qu2024Recursive, Kumar2025Training} whilst others have demonstrated inference efficiency gains by decomposing reasoning into a multi-turn process \citep{Sie2025Interleaved, Zheng2025Done}. \citet{Shen2025Thinking} investigate scaling test-time interaction as a multi-turn analogue to generating longer reasoning traces, thus shifting focus towards agentic capabilities. 

Recent research has begun investigating multi-turn interaction dynamics within multi-agent LLM systems. Training LLMs in self-play on zero-sum games has been found to improve their reasoning abilities in other domains \citep{Liu2025SPIRAL}, suggesting that multi-turn interaction between models could serve as a promising training environment for learning transferable skills. \citet{Perez2025When} introduce the use of ``cultural attractors'' as a tool for evaluating multi-turn information transmission chains between LLMs and \citet{Vallinder2025Cultural} investigate the conditions under which cooperation emerges between LLM agents.

\subsection{Learning from Language Feedback}
\label{language_feedback}

A critical component of collaborative problem-solving is the ability to learn from contextual feedback. Recent work has explored various methods for enabling LLMs to leverage language feedback. \citet{Scheurer2023Training} demonstrate that models can be finetuned on a large-scale dataset of human-written instructions and corresponding critiques to improve their instruction-following capabilities. Similarly, other research has focused on inferring reward functions directly from natural language descriptions of desired behaviours \citep{Lin2022Inferring}. While these approaches successfully use language as a supervisory signal, the feedback is typically provided by humans and applied in an offline manner to improve single-turn generation. More recent work has begun to formalise this problem. \citet{Xu2025Provably} develop a theoretical framework for learning from language feedback, conceptualising it as a reward function that can be optimised. Their work provides theoretical guarantees but focuses on language feedback as an abstract reward signal rather than part of an interactive dialogue.

\subsection{Social Reinforcement Learning}
\label{social_rl}

Our work is conceptually grounded in social and cultural learning (i.e., guided participation with more knowledgeable others) \citep{Vygotski1929The}, particularly as explored within multi-agent reinforcement learning (MARL). Research in this area has shown that learning from other agents in a shared environment can emerge without explicit rewards for doing so \citep{Borsa2019Observational, Ndousse2021Emergent}. Expanding on this, \citet{Cook2024Artificial} explore how knowledge and skills can be transmitted and accumulated across generations of agents --- a process analogous to human cultural evolution. While prior work has focused on agents in embodied or grid-world environments, we build on these principles by structuring LLM finetuning tasks as pedagogical social interactions. 

\section{Social Meta-Learning for Language Models}
\label{sec:SML}

\subsection{Problem Formulation}

To formally model the interactive process of learning from language feedback, we cast it as a partially observable Markov decision process (POMDP) $\langle \mathcal{S}, \mathcal{A}, O, \mathcal{O}, T, R, \gamma \rangle$, where the student model, parameterised by $\theta_S$, acts as the agent learning a policy $\pi_S$. A state $s_t \in \mathcal{S}$ at turn $t$ comprises of the public conversational history and the teacher's private knowledge, $s_t = (c_t, k_t)$, where $c_t = (u_1^T, u_1^S, u_2^T, u_2^S, \dots, u_t^T)$ is the conversation history and $k_t$ is the teacher's private knowledge, which encapsulates all information available to the teacher but not the student. In cases where the problem presented to the student is initially underspecified, $k_t$ can represent a complete, unambiguous problem specification. It can also represent the teacher's parametric knowledge in cases where the teacher and student are distinct models, or the knowledge held by a human user during live deployment. We consider settings where $k_t$ is static (e.g., a ground truth solution) and where it depends on the student utterances (e.g., the outputs of a verifier)\footnote{There are a number of interesting scenarios featuring non-stationarity in the task goal arising from a dynamic $k_t$, such as interactive recommendation, tutoring, or co-creation systems, which we leave to future work.}.

\subsection{Learning from Language Feedback (Inner Loop)}

The student's action space $\mathcal{A}$ consists of generating a natural language utterance. At each turn $t$, the student receives an observation $o_t \in \mathcal{O}$ corresponding to the conversation history, $o_t = c_t$. The observation function $O(s_t) = P(o_t | s_t)$ is therefore deterministic, as the student always observes the conversational part of the state, but remains unaware of the teacher's private knowledge $k_t$. The transition function $T(s_{t+1} | s_t, a_t)$ is implicitly defined by the teacher's policy $\pi_T$. The student's utterance $u_t^S = a_t$ is appended to the history, $c'_t = (c_t, u_t^S)$, and a response is generated by the teacher's policy, $u_{t+1}^T \sim \pi_T(\cdot | c'_t, k_t)$, which is conditioned on the updated history and its private knowledge. The final history for the next state is then $c_{t+1} = (c'_t, u_{t+1}^T)$. The environment's evolution thus depends on the hidden information revealed through the teacher's utterances. The reward function $R(s_t, a_t)$ can encode sparse, conversation-level rewards, such as final answer correctness scores, and/or dense, turn-level rewards, which could leverage the teacher as a judge of progress. We focus exclusively on sparse rewards, leaving exploration of turn-level rewards to future work.

\subsection{Meta-Training (Outer Loop)}
\label{meta_training}

During meta-training, we optimise the student model's parameters $\theta_S$ to produce a policy $\pi_{\theta_S}$ that is adept at in-context learning from the teacher's feedback. To facilitate this, we ground our training in verifiable domains, such as math and code, where performance can be objectively measured by whether a valid solution is generated by the student within the conversation. Each episode (i.e., dialogue) is a trajectory $\tau = (s_1, a_1, s_2, a_2, \dots, s_N, a_N)$ generated by the interaction between the student and a fixed teacher policy $\pi_T$. Episodes thus proceed as follows:

\begin{enumerate}
    \item \textbf{Initialisation:} A problem is sampled from $\mathcal{D}_\text{prompt}$. This forms the teacher's initial utterance $u_1^T$ and its private knowledge $k_t$. The initial state is therefore $s_1 = ((u_1^T), k_1)$. When $k_t$ represents the outputs of a verifier rather than a ground truth solution, $k_1$ is empty.
    \item \textbf{Rollout:} For a maximum of $N$ turns, the student and teacher interact. At each turn $t$, the student observes the history $o_t = c_t$ and generates an action $a_t = u_t^S \sim \pi_{\theta_S}(\cdot | c_t)$. The teacher then responds with $u_{t+1}^T \sim \pi_T(\cdot | (c_t, u_t^S), k_t)$, resulting in the next state $s_{t+1}$.
    \item \textbf{Termination \& Reward:} The dialogue terminates when the student produces a valid solution or the maximum turn limit $N$ is reached. A sparse, binary reward is then assigned to the entire trajectory, $G(\tau) = R(s_n, a_n)$.
\end{enumerate}


\paragraph{Offline RL.} Following the process described in Section \ref{prelims}, we first generate a dataset of conversational rollouts $\{ \tau_i \}$ using the initial student policy. We then filter this dataset to retain only the successful trajectories where $G(\tau_i) = 1$. The student's parameters $\theta_S$ are then updated via SFT on the student's turns from these successful dialogues\footnote{This offline RL approach is hereby referred to as SFT for simplicity.}. 

\paragraph{Online RL.} In the online setting, we employ GRPO. For each initial problem, we generate a group of $g$ conversational trajectories $\{ \tau_k \}_{k=1}^g$. Each trajectory is treated as a single sample and receives a trajectory-level reward $r_k = G(\tau_k)$. The advantages $A_k$ are calculated by normalising the rewards across the group and the policy is updated using the objective in Equation \ref{eq:grpo}. This directly reinforces the conversational strategies that lead to successful outcomes.

\subsection{Learning to Enquire with Q-Priming}
\label{sec:qpriming}

SML alone does not explicitly promote question asking (see Section \ref{seq:behaviour}), a desirable behaviour for navigating ambiguity. Motivated by the fact that models exhibiting certain cognitive behaviours have been found to better bootstrap their reasoning abilities during downstream finetuning \citep{Ghandi2025Cognitive}, we introduce \textbf{Q-priming}, a preliminary SFT stage to explicitly teach this skill. For this stage, we generate an SFT dataset by synthetically injecting questions into dialogues. During data generation, if a student's turn is incorrect, we replace its response with a generated question. The probability of this injection decays exponentially over turns (Equation \ref{eq:prob}), which encourages exploratory behaviour early in the conversation.

\begin{equation}
    P_Q(t) = 0.75^t \, \mathbb{I}[R(s_t, a_t) = 0]
    \label{eq:prob}
\end{equation}

To ensure the injected questions are useful, we again exploit the information asymmetry of our setup. We generate each question by providing the student model with its prior attempt and the teacher's private knowledge (e.g., the ground truth solution), then prompting it to formulate an informative query. This creates a dataset containing examples of effective, information-seeking questions.

\section{Experimental Setup}
\label{setup}

Our experiments are designed to investigate the efficacy and generalisability of SML. Specifically, we investigate the following research questions:

\begin{enumerate}
\item \textbf{RQ1. In-Domain Performance}: Does SML improve a model's ability to learn from language feedback on tasks from the training domain more than single-turn finetuning?
\item \textbf{RQ2. Online vs. Offline Training}: Is online or offline training more effective for SML?
\item \textbf{RQ3. Cross-Domain Generalisation}: Can the ability to learn from language feedback, acquired via SML in one domain, transfer to a different domain?
\item \textbf{RQ4. Feedback Generalisation}: Can a model trained on interactions with corrective feedback generalise to different forms of language feedback, such as incremental problem specification, and can it generalise to longer conversations than were used for training?
\item \textbf{RQ5. Behavioural Adaptation}: Does SML encourage more effective conversational strategies when faced with ambiguity, such as asking clarifying questions?
\end{enumerate}

\subsection{Datasets and Evaluation Protocol}

\paragraph{Training Data.}
We use two datasets to train our models, corresponding to our two domains of interest: mathematics and coding. For math training, we use a training set of 2,000 problems from the Omni-MATH \citep{Gao2025Omni}, a large and diverse collection of Olympiad-level competition math problems. For code training, we use OpenCodeInstruct \citep{Ahmad2025OpenCodeInstruct}, a large-scale dataset of code generation problems with test cases.

\paragraph{Evaluation Tasks.}
We evaluate models in three distinct settings designed to probe different facets of their interactive learning capabilities.

\textbf{In-Domain: Learning from Language Feedback for Math.} To assess in-domain performance, we use a held-out test set of 500 problems from Omni-MATH. The evaluation mirrors the training setup: the student model engages in a multi-turn dialogue with a teacher model to solve a math problem. This setting is analogous to the one proposed by the MINT benchmark \citep{Wang2024MINT}, where success is measured by the model's ability to reach a correct solution through conversation with a teacher that provides guidance based on a ground truth solution. We use Omni-MATH rather than the math problems in MINT, as single-turn performance on Omni-MATH is less saturated.

\textbf{Cross-Domain Transfer.} We test the ability of models to generalise the skill of learning from feedback across domains in two directions.

\begin{itemize}
    \item \textbf{Math-to-Code:} To test transfer from math to code, we evaluate models trained on Omni-MATH on code generation tasks from LiveCodeBench \citep{Jain2025LiveCodeBench}. Here, the teacher's private knowledge is comprised of public test cases and their outputs. The teacher provides feedback based on information revealed by test outputs, simulating a common developer workflow.
    \item \textbf{Code-to-Math:} Conversely, to test transfer from code to math, we evaluate models trained on $2,000$ problems from OpenCodeInstruct, again using test outputs as the teacher's private knowledge, on the held-out Omni-MATH test set. The interactive evaluation protocol for this task is identical to the in-domain setting.
\end{itemize}

\textbf{Handling Ambiguity: Sharded Problems.} We evaluate models' ability to handle underspecified problems using the math and code subtasks from the \textbf{Lost-in-Conversation} benchmark \citep{Laban2025LLMs}. In this setting, the problem statement is ``sharded'' and revealed incrementally over several turns. The teacher simulates a ``lazy user,'' providing the next piece of information (i.e., shard) in each turn. The dialogue terminates if the student provides a correct solution or if all shards have been revealed and the student has failed to solve the problem.

\subsection{Behavioural Analysis}
To understand how SML alters a model's conversational behaviour, we conduct a behavioural analysis following the protocol defined by the Lost-in-Conversation benchmark. This involves using a strong model-as-judge (we use \texttt{Gemini-2.5-Pro} \citep{Comanici2025Gemini}) to classify each student turn into a predefined set of turn types (e.g., answer attempt, clarification question, discussion). We then analyse the frequency of these conversational actions to see how the student's approach to soliciting and integrating language feedback changes as a result of SML finetuning.

\subsection{Baselines}
We compare SML against three baselines that do not involve multi-turn interaction during training:
\begin{itemize}
\item \textbf{Single-turn SFT}: A standard SFT baseline trained on (problem, solution) pairs.
\item \textbf{Single-turn RL}: A baseline trained with the same GRPO algorithm on single-turn generations. This controls for the effect of the RL algorithm. 
\item \textbf{Scaled group size}: To control for the increased number of generation steps in SML, we also evaluate a variant of the single-turn RL baseline where we use a $4\times$ larger group size ($g=32$) to provide it with a comparable generation budget. This baseline therefore involves training on a larger number of trajectories, but with fewer tokens per trajectory.
\end{itemize}
We also directly compare the performance of our offline and online implementations of SML, and investigate the impact of using a stronger model as the teacher.

\subsection{Implementation Details}
For all online RL experiments (both multi-turn SML and the single-turn baseline), we use GRPO with a group size of $g=8$ and no KL regularisation ($\beta=0$). We apply reward discounting over the conversational turns with a discount factor of $\gamma=0.7$. For the SML rollouts, we use a maximum of $N=4$ turns. However, we evaluate on Omni-MATH and LiveCodeBench with $N=10$ turns to investigate generalisation to longer conversations. All finetuning runs are performed for a single epoch over three seeds and $95\%$ confidence intervals are reported for each evaluation of a finetuned model. We use \texttt{Gemma-3-12B-IT} \citep{Kamath2025Gemma} as our primary base model and repeat core experiments with \texttt{Qwen3-8B} \citep{Yang2025Qwen3} in Appendix \ref{ap:qwen}. 

\section{Results}
\label{results}

\subsection{RQ1 \& RQ2: In-Domain Performance \& Online vs. Offline Training}
\label{sec:results_1_2}

\begin{wrapfigure}{l}{8cm}
\vspace{-0.75cm}
\begin{center}
\includegraphics[width=7.5cm, height=6cm]{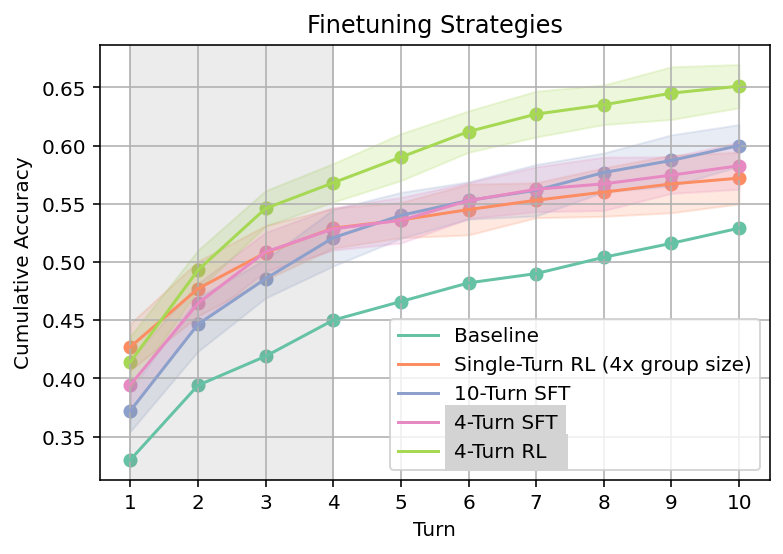}
\end{center}
\vspace{-0.7cm}
\caption{Comparing different multi-turn and single-turn finetuning strategies on Omni-MATH.}
\vspace{-0.75cm}
\label{fig:id}
\end{wrapfigure}

Figure \ref{fig:id} presents a comparison between different multi-turn finetuning strategies for SML on Omni-MATH. We observe that online RL is more effective at generalising to test problems, which is in line with a growing body of prior work comparing generalisation from RL and SFT/ imitation learning \citep{Ostrovski2021The, Kirk2024Understanding, Chu2025SFT, Cook2025Programming, Kumar2025Training}. Interestingly, training on conversations with a maximum of 4 turns enables continuing to learn from language feedback for up to 10 turns at test time. 4-turn and 10-turn SFT perform comparably when evaluated with $N=10$, indicating that there are diminishing returns from scaling the number of interaction turns used in SFT on Omni-MATH. We also include the single-turn RL baseline with $4 \times$ larger GRPO group size. This performs comparably on the first turn, but the model trained with 4-turn RL establishes a widening performance gap over longer conversations. This ability to improve from conversational feedback is crucial for human-AI interaction, making dialogues less brittle and more productive when a user provides corrections.

\begin{figure}[t]
\begin{center}
\includegraphics[width=12cm, height=5cm]{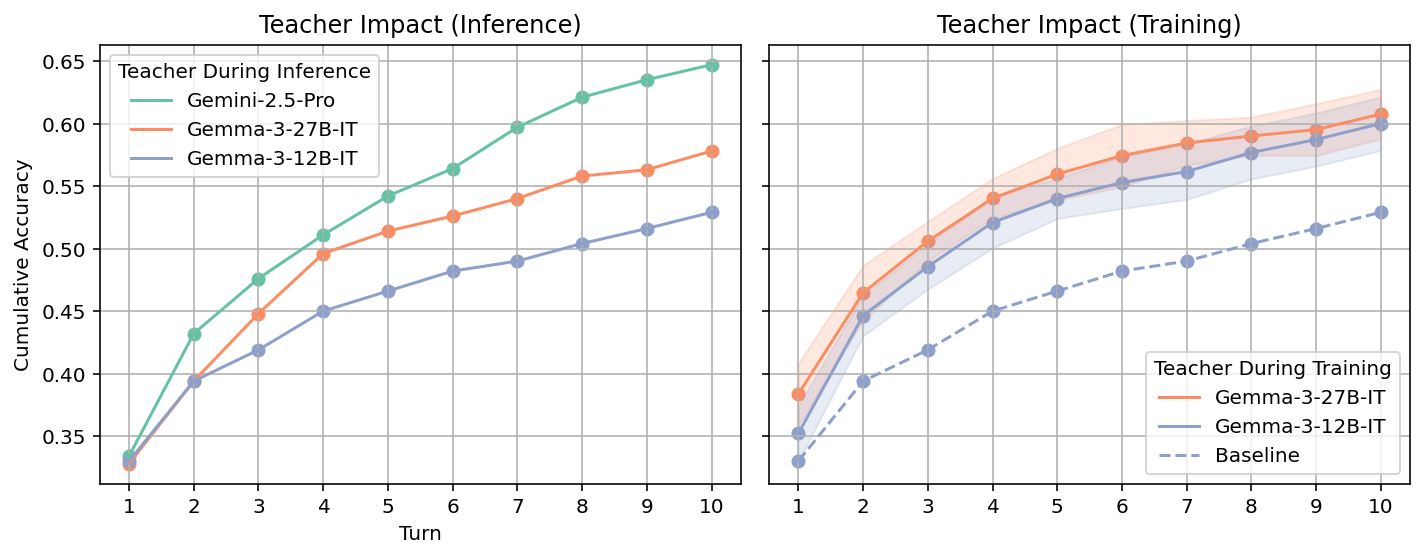}
\end{center}
\vspace{-0.3cm}
\caption{Evaluating the impact of using stronger models as teachers on Omni-MATH. In the right plot, the same teacher (\texttt{Gemma-3-12B-IT}) is used for training each student at test time.}
\label{fig:teachers}
\end{figure}

In Figure \ref{fig:teachers} (left), we show the impact of using a stronger model as the teacher at inference time, revealing that more powerful teachers provide feedback that is more informative to the student. Here, the student being evaluated is \texttt{Gemma-3-12B-IT} without any further finetuning. Figure \ref{fig:teachers} (right) shows the impact of using a stronger model as teacher during training. At test time, each student model is evaluated using \texttt{Gemma-3-12B-IT} as the teacher. The teacher model used for training has much less impact than during inference, indicating that a stronger teacher model is unnecessary for SML training. This suggests that the skill of learning from feedback can be acquired even from a peer-level conversational partner.

\begin{wrapfigure}{r}{8cm}
\vspace{-2.5cm}
\begin{center}
\includegraphics[width=7.5cm, height=6cm]{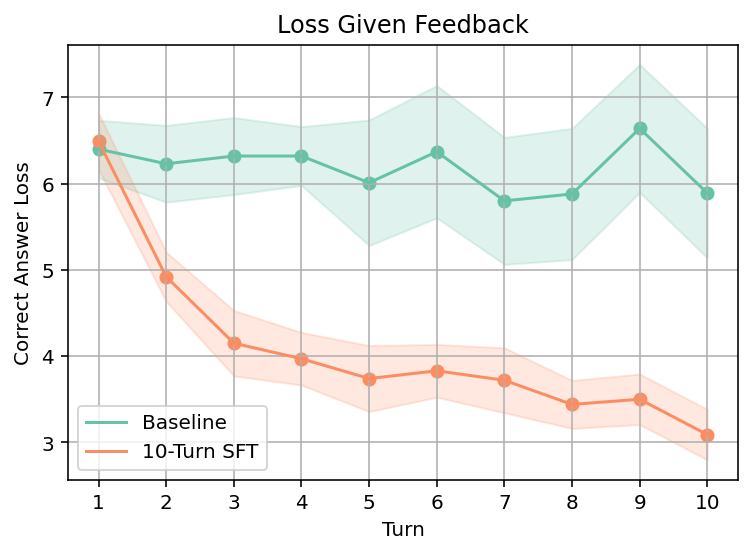}
\end{center}
\vspace{-0.7cm}
\caption{Average loss on the correct answer across conversational turns for Omni-MATH.}
\vspace{-1cm}
\label{fig:loss}
\end{wrapfigure}

To illustrate changes to in-context learning dynamics following SML, we compute the loss on the correct answer following each teacher turn in a conversation for the base model and a model that was trained on successful 10-turn conversations. These results, averaged across the Omni-MATH test split, are shown in Figure \ref{fig:loss}. The model trained via SML exhibits a clear loss reduction as conversations progress, whilst the base model shows no clear trend. This clearly demonstrates that SML teaches models \emph{how} to learn from language feedback.

\subsection{RQ3: Cross-Domain Transfer}
\label{sec:results_3}

\begin{figure}[t]
\begin{center}
\includegraphics[width=12cm, height=5cm]{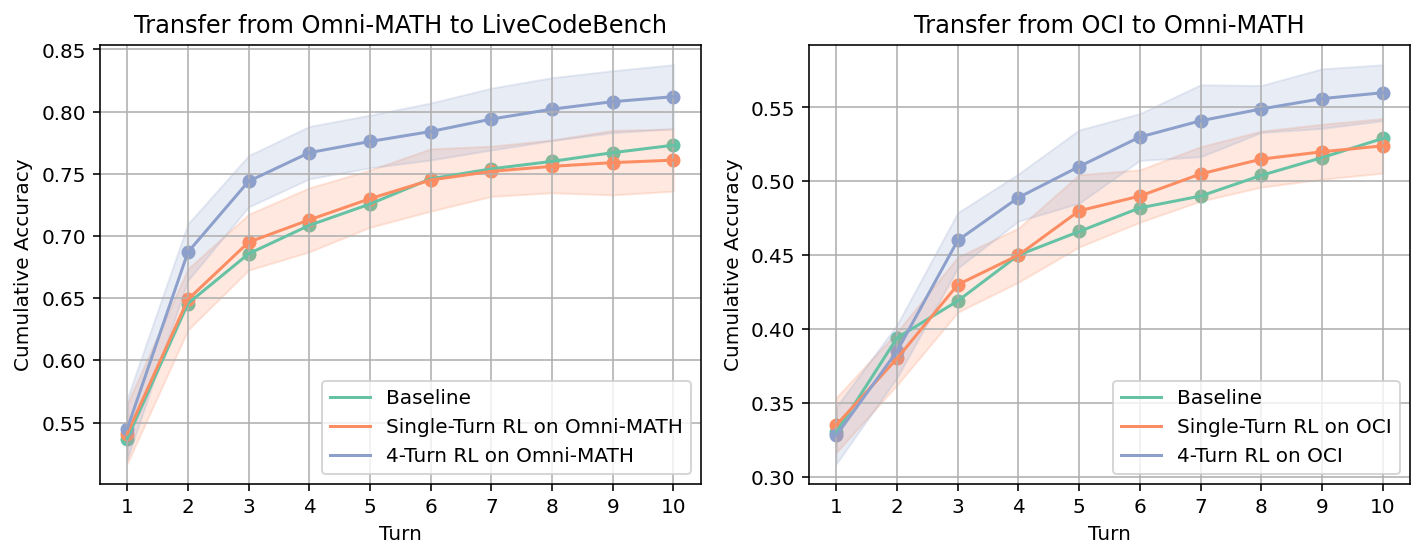}
\end{center}
\vspace{-0.3cm}
\caption{Evaluating transfer of the ability to learn from language feedback between math and code domains. Left: Training on Omni-MATH and evaluating on LiveCodeBench; Right: Training on OpenCodeInstruct and evaluating on Omni-MATH.}
\label{fig:transfer}
\end{figure}

In Figure \ref{fig:transfer}, we compare the generalisation of models trained on Omni-MATH to LiveCodeBench. For this task, rather than a ground truth solution, the teacher's private knowledge consists of the outputs of the public test cases, simulating a common developer workflow where a user provides feedback based on test results. Both finetuning approaches yield comparable single-turn performance to the baseline, but SML (4-turn RL) leads to a notable performance gain over a multi-turn conversations. This demonstrates that the ability to learn from language feedback is a general skill that transfers across domains, which could enable human users to collaborate more effectively with chat-based assistants in a variety of settings.

\subsection{RQ4: Feedback Generalisation}
\label{sec:results_4}

We report the Lost-in-Conversation (math \& code) performance following different approaches to finetuning on Omni-MATH in Table \ref{tab:lost}. First, baseline \texttt{Gemma-3-12B-IT} results reflect the central finding of \citet{Laban2025LLMs}; performance on the sharded tasks is much lower than on the fully specified tasks. Second, while single-turn RL performs the best on the full tasks, SML (MT-RL) leads to considerably better generalisation to the sharded tasks. Multi-turn SFT also outperforms single-turn baselines in this setting. These results show that training on dialogues with explicit, corrective feedback helps models adapt to a communication style that closely reflects how users often engage with chat-based assistants, where information is provided piece by piece. This reduces the burden on the user to craft a perfect, exhaustive initial prompt. These results also demonstrate another instance of cross-domain transfer from math to code.

\begin{table}[h!]
\centering
\begin{tabular}{*7l} \toprule
        &  & Baseline                       & ST-SFT                         & ST-RL                                                    & MT-SFT                         & MT-RL                                                    \\ \midrule
\multirow{2}{*}{Math} & Full & $92.0$ & $92.5 \pm 2$ & $\mathbf{94.4} \pm 1$ & $93.2 \pm 2$ & $94.1 \pm 2$                           \\ 
 & Sharded & $62.7$ & $63.3 \pm 1$ & $63.7 \pm 2$                           & $67.1 \pm 2$ & $\mathbf{69.9} \pm 1$ \\ 
\midrule
\multirow{2}{*}{Code} & Full    & 76.3 & $76.7 \pm 3$ & $\mathbf{77.1} \pm 2$ & $75.8 \pm 1$ & $76.2 \pm 2$                          \\ 
& Sharded & 51.5 & $48.9 \pm 3$ & $51.6 \pm 2$                           & $53.0 \pm 3$ & $\mathbf{54.8} \pm 2$ \\ \bottomrule
\end{tabular}
\caption{Evaluating transfer to the math and code subtasks of the Lost-in-Conversation benchmark. Training was done on Omni-MATH.}
\label{tab:lost}
\end{table}

\begin{figure}[t]
\begin{center}
\includegraphics[width=12cm, height=5cm]{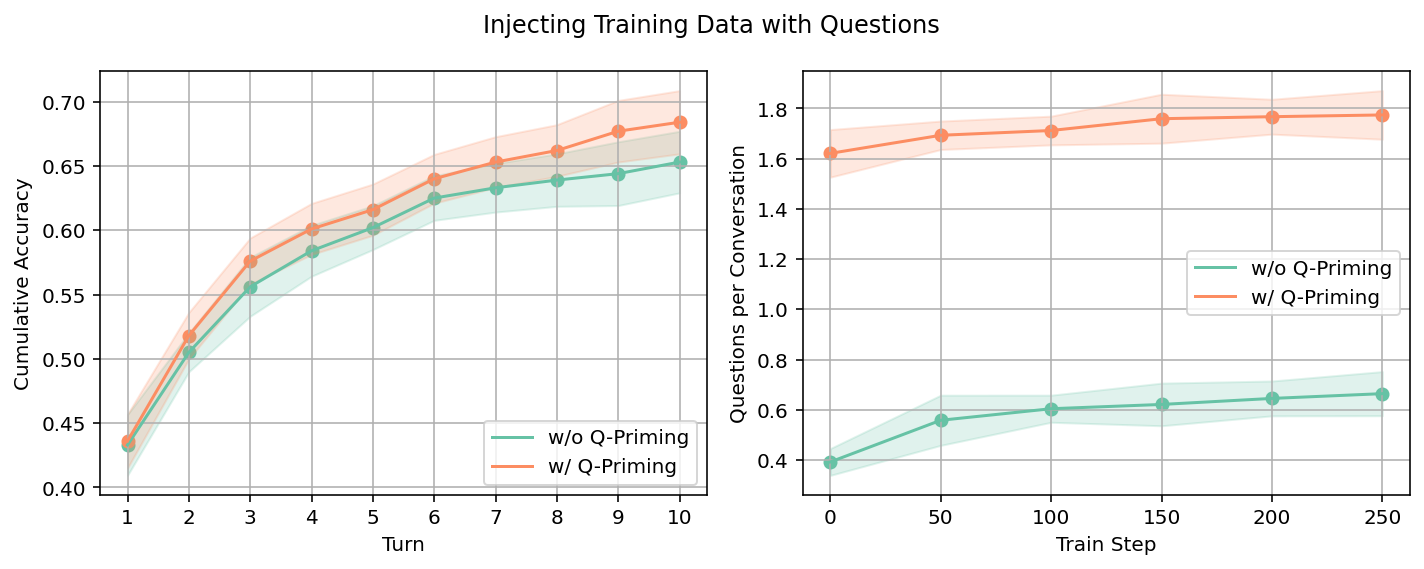}
\end{center}
\vspace{-0.3cm}
\caption{Evaluating the impact of Q-priming on multi-turn performance (left) and the rate of question asking per conversation (right) for Omni-MATH.}
\label{fig:questioning}
\end{figure}

\subsection{RQ5: Behavioural Adaptation}
\label{seq:behaviour}

To compare behavioural adaptation with and without Q-priming, we perform SML in two stages: first SFT, then RL. With Q-priming, the SFT stage uses a dataset containing injected questions (see Section \ref{sec:qpriming}). Figure \ref{fig:questioning} shows that Q-priming successfully instills and maintains question-asking behaviour during subsequent RL training. Without it, the model only slowly discovers question-asking as a useful strategy. Furthermore, Q-priming leads to performance gains relative to using an SFT dataset without injected questions, highlighting the importance of exploratory behaviour.

\begin{figure}[h!]
\begin{center}
\includegraphics[width=9.5cm, height=5cm]{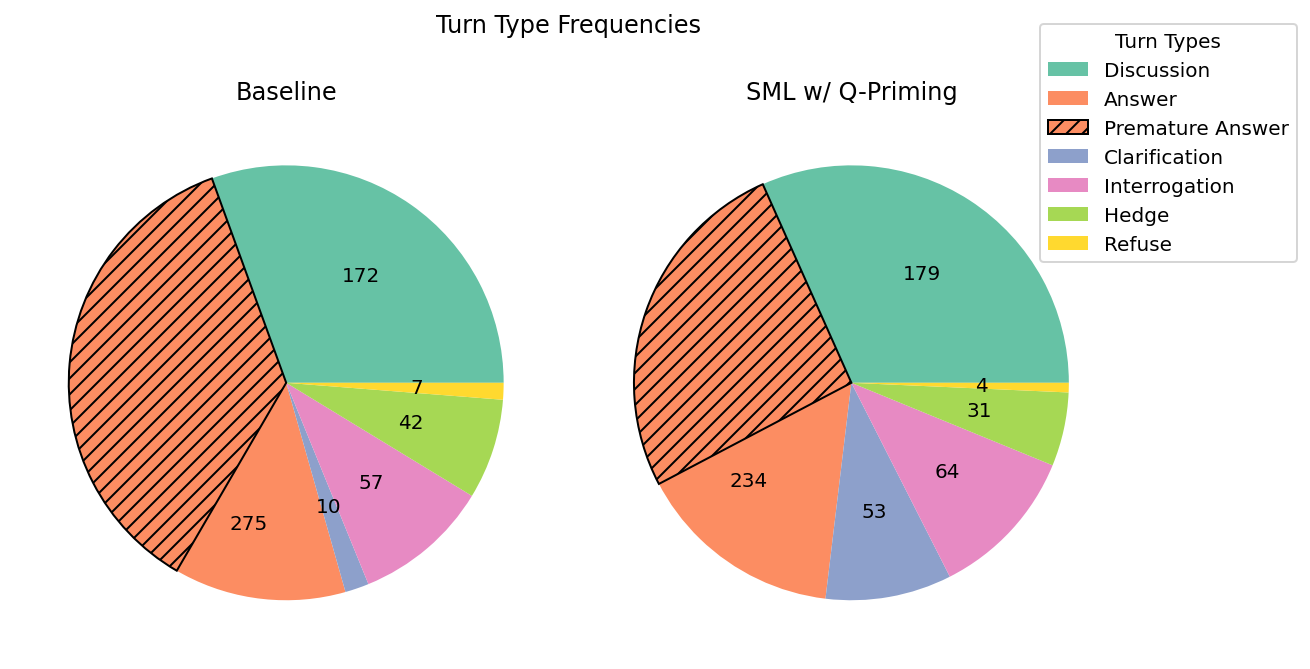}
\end{center}
\caption{Comparing the conversational behaviours exhibited by models on Lost-in-Conversation before and after SML with Q-priming.}
\vspace{-0.1cm}
\label{fig:actions}
\end{figure}

Another practical benefit of this behavioural shift is shown in Figure \ref{fig:actions}. On the ambiguous Lost-in-Conversation benchmark, the SML model with Q-priming makes far fewer premature answer attempts. Instead, it is over 5 times more likely to ask a clarifying question. This change from presumptive guessing to proactive enquiry marks a significant step towards more collaborative AI. For the user, this means a more efficient dialogue, as the model helps clarify ambiguity rather than confidently providing incorrect answers based on assumptions.

\section{Conclusion}
\label{conclusion}

In this work, we introduce a finetuning methodology, inspired by the phenomenon of social meta-learning (SML) in humans, that trains LLMs to learn how to learn from others in conversation. By converting static problems into interactive pedagogical dialogues, we show that models can be taught to use language feedback to solve problems they are unable to solve in a single turn. Our results demonstrate that this learned ability is highly generalisable: SML on math problems improves a model's capacity to learn from feedback on coding tasks and to handle ambiguity in underspecified problems. Furthermore, by introducing a Q-priming stage, we can elicit exploration within conversation, encouraging behaviours such as asking clarifying questions instead of making premature answer attempts. This work presents a scalable path toward developing more collaborative and human-compatible AI by reframing static tasks as interactive learning opportunities.

\paragraph{Limitations \& Future Work.}

Our approach is demonstrated in verifiable domains (math and code) with sparse, conversation-level rewards. Future work should extend SML to more subjective, open-ended tasks, which will require use of more nuanced reward models. Exploring denser, turn-level rewards could also improve sample efficiency. Finally, our experiments assume a static underlying task goal; applying SML to dynamic settings where user intent, knowledge, or preference evolves during conversation remains an important direction for future research.

\bibliography{main}

@article{Allen2017Social,
    author = {Jedediah W. P. Allen and Hande Ilgaz},
    journal = {Frontiers in Artificial Intelligence and Applications},
    pages = {89--113},
    title = {Social meta-learning: Learning how to make use of others as a resource for further learning},
    volume = {18},
    publisher = {Springer International Publishing},
    year = {2017}
}

@article{Xu2025Provably,
  author       = {Wanqiao Xu and
                  Allen Nie and
                  Ruijie Zheng and
                  Aditya Modi and
                  Adith Swaminathan and
                  Ching{-}An Cheng},
  title        = {Provably Learning from Language Feedback},
  journal      = {arXiv preprint},
  year         = {2025},
  url          = {https://doi.org/10.48550/arXiv.2506.10341},
  eprinttype    = {arXiv},
  eprint       = {2506.10341}
}

@article{Liu2025SPIRAL,
  author       = {Bo Liu and
                  Leon Guertler and
                  Simon Yu and
                  Zichen Liu and
                  Penghui Qi and
                  Daniel Balcells and
                  Mickel Liu and
                  Cheston Tan and
                  Weiyan Shi and
                  Min Lin and
                  Wee Sun Lee and
                  Natasha Jaques},
  title        = {{SPIRAL:} Self-Play on Zero-Sum Games Incentivizes Reasoning via Multi-Agent
                  Multi-Turn Reinforcement Learning},
  journal      = {arXiv preprint},
  year         = {2025},
  url          = {https://doi.org/10.48550/arXiv.2506.24119},
  eprinttype    = {arXiv},
  eprint       = {2506.24119}
}

@article{Li2025Beyond,
  author       = {Yubo Li and
                  Xiaobin Shen and
                  Xinyu Yao and
                  Xueying Ding and
                  Yidi Miao and
                  Ramayya Krishnan and
                  Rema Padman},
  title        = {Beyond Single-Turn: {A} Survey on Multi-Turn Interactions with Large
                  Language Models},
  journal      = {arXiv preprint},
  year         = {2025},
  url          = {https://doi.org/10.48550/arXiv.2504.04717},
  eprinttype    = {arXiv},
  eprint       = {2504.04717}
}

@article{Scheurer2023Training,
  author       = {J{\'{e}}r{\'{e}}my Scheurer and
                  Jon Ander Campos and
                  Tomasz Korbak and
                  Jun Shern Chan and
                  Angelica Chen and
                  Kyunghyun Cho and
                  Ethan Perez},
  title        = {Training Language Models with Language Feedback at Scale},
  journal      = {arXiv preprint},
  volume       = {abs/2303.16755},
  year         = {2023},
  url          = {https://doi.org/10.48550/arXiv.2303.16755},
  eprinttype    = {arXiv},
  eprint       = {2303.16755}
}

@article{Laban2025LLMs,
  author       = {Philippe Laban and
                  Hiroaki Hayashi and
                  Yingbo Zhou and
                  Jennifer Neville},
  title        = {LLMs Get Lost In Multi-Turn Conversation},
  journal      = {arXiv preprint},
  year         = {2025},
  url          = {https://doi.org/10.48550/arXiv.2505.06120},
  eprinttype    = {arXiv},
  eprint       = {2505.06120}
}

@article{Zhou2025SWEET,
  author       = {Yifei Zhou and
                  Song Jiang and
                  Yuandong Tian and
                  Jason Weston and
                  Sergey Levine and
                  Sainbayar Sukhbaatar and
                  Xian Li},
  title        = {{SWEET-RL:} Training Multi-Turn {LLM} Agents on Collaborative Reasoning
                  Tasks},
  journal      = {arXiv preprint},
  year         = {2025},
  url          = {https://doi.org/10.48550/arXiv.2503.15478},
  eprinttype    = {arXiv},
  eprint       = {2503.15478},
}

@article{Li2025Quest,
  author       = {Belinda Z. Li and
                  Been Kim and
                  Zi Wang},
  title        = {QuestBench: Can LLMs ask the right question to acquire information
                  in reasoning tasks?},
  journal      = {arXiv preprint},
  year         = {2025},
  url          = {https://doi.org/10.48550/arXiv.2503.22674},
  eprinttype    = {arXiv},
  eprint       = {2503.22674},
}

@inproceedings{Kumar2025Training,
  author       = {Aviral Kumar and
                  Vincent Zhuang and
                  Rishabh Agarwal and
                  Yi Su and
                  John D. Co{-}Reyes and
                  Avi Singh and
                  Kate Baumli and
                  Shariq Iqbal and
                  Colton Bishop and
                  Rebecca Roelofs and
                  Lei M. Zhang and
                  Kay McKinney and
                  Disha Shrivastava and
                  Cosmin Paduraru and
                  George Tucker and
                  Doina Precup and
                  Feryal M. P. Behbahani and
                  Aleksandra Faust},
  title        = {Training Language Models to Self-Correct via Reinforcement Learning},
  booktitle    = {The Thirteenth International Conference on Learning Representations,
                  {ICLR} 2025, Singapore, April 24-28, 2025},
  year         = {2025},
  url          = {https://openreview.net/forum?id=CjwERcAU7w},
}

@article{Ghandi2025Cognitive,
  author       = {Kanishk Gandhi and
                  Ayush Chakravarthy and
                  Anikait Singh and
                  Nathan Lile and
                  Noah D. Goodman},
  title        = {Cognitive Behaviors that Enable Self-Improving Reasoners, or, Four
                  Habits of Highly Effective STaRs},
  journal      = {arXiv preprint},
  year         = {2025},
  url          = {https://doi.org/10.48550/arXiv.2503.01307},
  eprinttype    = {arXiv},
  eprint       = {2503.01307}
}

@article{Shen2025Thinking,
  author       = {Junhong Shen and
                 Hoa Bai and
                 Lunjun Zhang and
                 Yifei Zhou and
                 Amrith Setlur and
                 Shengbang Tong and
                 Diego Caples and
                 Nan Jiang and
                 Tong Zhang and
                 Ameet Talwalkar and
                 Aviral Kumar},
  title        = {Thiking vs. Doing: Agents that Reason by Scaling Test-Time Interaction},
  journal      = {arXiv preprint},
  year         = {2025},
  url          = {https://www.arxiv.org/pdf/2506.07976},
  eprinttype    = {arXiv},
  eprint       = {2506.07976}
}

@article{Sie2025Interleaved,
  author       = {Roy Xie and
                 David Qiu and
                 Deepak Gopinath and
                 Dong Lin and
                 Yanchao Sun and
                 Chong Wang and 
                 Saloni Potdar and
                 Bhuwan Dhingra},
  title        = {Interleaved Reasoning for Large Language Models via Reinforcement Learning},
  journal      = {arXiv preprint},
  year         = {2025},
  url          = {https://arxiv.org/pdf/2505.19640},
  eprinttype    = {arXiv},
  eprint       = {2505.19640}
}

@article{Zheng2025Done,
  author       = {Zihao Zheng and
                 Xuyao Huang and
                 Boxiu Li and
                 Hoa Zhang and
                 Zhijie Deng},
  title        = {Done Is Better than Perfect: Unlocking Efficient Reasoning by Structured Multi-Turn Decomposition},
  journal      = {arXiv preprint},
  year         = {2025},
  url          = {https://arxiv.org/pdf/2505.19788},
  eprinttype    = {arXiv},
  eprint       = {2505.19788}
}

@article{Shao2024DeepSeekMath,
  author       = {Zhihong Shao and
                  Peiyi Wang and
                  Qihao Zhu and
                  Runxin Xu and
                  Junxiao Song and
                  Mingchuan Zhang and
                  Y. K. Li and
                  Y. Wu and
                  Daya Guo},
  title        = {DeepSeekMath: Pushing the Limits of Mathematical Reasoning in Open
                  Language Models},
  journal      = {arXiv preprint},
  year         = {2024},
  url          = {https://doi.org/10.48550/arXiv.2402.03300},
  eprinttype    = {arXiv},
  eprint       = {2402.03300}
}

@inproceedings{Gao2025Omni,
  author       = {Bofei Gao and
                  Feifan Song and
                  Zhe Yang and
                  Zefan Cai and
                  Yibo Miao and
                  Qingxiu Dong and
                  Lei Li and
                  Chenghao Ma and
                  Liang Chen and
                  Runxin Xu and
                  Zhengyang Tang and
                  Benyou Wang and
                  Daoguang Zan and
                  Shanghaoran Quan and
                  Ge Zhang and
                  Lei Sha and
                  Yichang Zhang and
                  Xuancheng Ren and
                  Tianyu Liu and
                  Baobao Chang},
  title        = {Omni-MATH: {A} Universal Olympiad Level Mathematic Benchmark for Large
                  Language Models},
  booktitle    = {The Thirteenth International Conference on Learning Representations,
                  {ICLR} 2025, Singapore, April 24-28, 2025},
  year         = {2025},
  url          = {https://openreview.net/forum?id=yaqPf0KAlN}
}

@inproceedings{Jain2025LiveCodeBench,
  author       = {Naman Jain and
                  King Han and
                  Alex Gu and
                  Wen{-}Ding Li and
                  Fanjia Yan and
                  Tianjun Zhang and
                  Sida Wang and
                  Armando Solar{-}Lezama and
                  Koushik Sen and
                  Ion Stoica},
  title        = {LiveCodeBench: Holistic and Contamination Free Evaluation of Large
                  Language Models for Code},
  booktitle    = {The Thirteenth International Conference on Learning Representations,
                  {ICLR} 2025, Singapore, April 24-28, 2025},
  year         = {2025},
  url          = {https://openreview.net/forum?id=chfJJYC3iL}
}

@inproceedings{Vallinder2025Cultural,
  author       = {Aron Vallinder and
                  Edward Hughes},
  editor       = {Sanmay Das and
                  Ann Now{\'{e}} and
                  Yevgeniy Vorobeychik},
  title        = {Cultural Evolution of Cooperation among {LLM} Agents},
  booktitle    = {Proceedings of the 24th International Conference on Autonomous Agents
                  and Multiagent Systems, {AAMAS} 2025, Detroit, MI, USA, May 19-23,
                  2025},
  pages        = {2771--2773},
  publisher    = {International Foundation for Autonomous Agents and Multiagent Systems
                  / {ACM}},
  year         = {2025},
  url          = {https://dl.acm.org/doi/10.5555/3709347.3744007}
}

@inproceedings{Perez2025When,
  author       = {J{\'{e}}r{\'{e}}my Perez and
                  Grgur Kovac and
                  Corentin L{\'{e}}ger and
                  C{\'{e}}dric Colas and
                  Gaia Molinaro and
                  Maxime Derex and
                  Pierre{-}Yves Oudeyer and
                  Cl{\'{e}}ment Moulin{-}Frier},
  title        = {When LLMs Play the Telephone Game: Cultural Attractors as Conceptual
                  Tools to Evaluate LLMs in Multi-turn Settings},
  booktitle    = {The Thirteenth International Conference on Learning Representations,
                  {ICLR} 2025, Singapore, April 24-28, 2025},
  year         = {2025},
  url          = {https://openreview.net/forum?id=fN8yLc3eA7}
}

@inproceedings{Cook2024Artificial,
  author       = {Jonathan Cook and
                  Chris Lu and
                  Edward Hughes and
                  Joel Z. Leibo and
                  Jakob N. Foerster},
  title        = {Artificial Generational Intelligence: Cultural Accumulation in Reinforcement
                  Learning},
  booktitle    = {Advances in Neural Information Processing Systems 38: Annual Conference
                  on Neural Information Processing Systems 2024, NeurIPS 2024, Vancouver,
                  BC, Canada, December 10 - 15, 2024},
  year         = {2024},
  url          = {http://papers.nips.cc/paper\_files/paper/2024/hash/6df3a719d99bd2479c04114d357003d0-Abstract-Conference.html}
}

@inproceedings{Ndousse2021Emergent,
  author       = {Kamal Ndousse and
                  Douglas Eck and
                  Sergey Levine and
                  Natasha Jaques},
  title        = {Emergent Social Learning via Multi-agent Reinforcement Learning},
  booktitle    = {Proceedings of the 38th International Conference on Machine Learning,
                  {ICML} 2021, 18-24 July 2021, Virtual Event},
  series       = {Proceedings of Machine Learning Research},
  volume       = {139},
  pages        = {7991--8004},
  publisher    = {{PMLR}},
  year         = {2021},
  url          = {http://proceedings.mlr.press/v139/ndousse21a.html}
}

@article{Yi2024Survey,
  author       = {Zihao Yi and
                  Jiarui Ouyang and
                  Yuwen Liu and
                  Tianhao Liao and
                  Zhe Xu and
                  Ying Shen},
  title        = {A Survey on Recent Advances in LLM-Based Multi-turn Dialogue Systems},
  journal      = {arXiv preprint},
  year         = {2024},
  url          = {https://doi.org/10.48550/arXiv.2402.18013},
  eprinttype    = {arXiv},
  eprint       = {2402.18013}
}

@inproceedings{Zhang2025Modeling,
  author       = {Michael Jq Zhang and
                  W. Bradley Knox and
                  Eunsol Choi},
  title        = {Modeling Future Conversation Turns to Teach LLMs to Ask Clarifying
                  Questions},
  booktitle    = {The Thirteenth International Conference on Learning Representations,
                  {ICLR} 2025, Singapore, April 24-28, 2025},
  year         = {2025},
  url          = {https://openreview.net/forum?id=cwuSAR7EKd}
}

@article{Andukuri2024STaR,
  author       = {Chinmaya Andukuri and
                  Jan{-}Philipp Fr{\"{a}}nken and
                  Tobias Gerstenberg and
                  Noah D. Goodman},
  title        = {STaR-GATE: Teaching Language Models to Ask Clarifying Questions},
  journal      = {arXiv preprint},
  year         = {2024},
  url          = {https://doi.org/10.48550/arXiv.2403.19154},
  eprinttype    = {arXiv},
  eprint       = {2403.19154}
}

@article{Chen2025Broaden,
  author       = {Zhiliang Chen and
                  Xinyuan Niu and
                  Chuan{-}Sheng Foo and
                  Bryan Kian Hsiang Low},
  title        = {Broaden your SCOPE! Efficient Multi-turn Conversation Planning for
                  LLMs using Semantic Space},
  journal      = {arXiv preprint},
  year         = {2025},
  url          = {https://doi.org/10.48550/arXiv.2503.11586},
  eprinttype    = {arXiv},
  eprint       = {2503.11586}
}

@article{Wu2025CollabLLM,
  author       = {Shirley Wu and
                  Michel Galley and
                  Baolin Peng and
                  Hao Cheng and
                  Gavin Li and
                  Yao Dou and
                  Weixin Cai and
                  James Zou and
                  Jure Leskovec and
                  Jianfeng Gao},
  title        = {CollabLLM: From Passive Responders to Active Collaborators},
  journal      = {arXiv preprint},
  year         = {2025},
  url          = {https://doi.org/10.48550/arXiv.2502.00640},
  eprinttype    = {arXiv},
  eprint       = {2502.00640}
}

@inproceedings{Qu2024Recursive,
  author       = {Yuxiao Qu and
                  Tianjun Zhang and
                  Naman Garg and
                  Aviral Kumar},
  title        = {Recursive Introspection: Teaching Language Model Agents How to Self-Improve},
  booktitle    = {Advances in Neural Information Processing Systems 38: Annual Conference
                  on Neural Information Processing Systems 2024, NeurIPS 2024, Vancouver,
                  BC, Canada, December 10 - 15, 2024},
  year         = {2024},
  url          = {http://papers.nips.cc/paper\_files/paper/2024/hash/639d992f819c2b40387d4d5170b8ffd7-Abstract-Conference.html}
}

@inproceedings{Zhou2024ArCHer,
  author       = {Yifei Zhou and
                  Andrea Zanette and
                  Jiayi Pan and
                  Sergey Levine and
                  Aviral Kumar},
  title        = {ArCHer: Training Language Model Agents via Hierarchical Multi-Turn
                  {RL}},
  booktitle    = {Forty-first International Conference on Machine Learning, {ICML} 2024,
                  Vienna, Austria, July 21-27, 2024},
  year         = {2024},
  url          = {https://openreview.net/forum?id=b6rA0kAHT1}
}

@inproceedings{Lin2022Inferring,
  author       = {Jessy Lin and
                  Daniel Fried and
                  Dan Klein and
                  Anca D. Dragan},
  title        = {Inferring Rewards from Language in Context},
  booktitle    = {Proceedings of the 60th Annual Meeting of the Association for Computational
                  Linguistics (Volume 1: Long Papers), {ACL} 2022, Dublin, Ireland,
                  May 22-27, 2022},
  pages        = {8546--8560},
  publisher    = {Association for Computational Linguistics},
  year         = {2022},
  url          = {https://doi.org/10.18653/v1/2022.acl-long.585}
}

@inproceedings{Wang2024MINT,
  author       = {Xingyao Wang and
                  Zihan Wang and
                  Jiateng Liu and
                  Yangyi Chen and
                  Lifan Yuan and
                  Hao Peng and
                  Heng Ji},
  title        = {{MINT:} Evaluating LLMs in Multi-turn Interaction with Tools and Language
                  Feedback},
  booktitle    = {The Twelfth International Conference on Learning Representations,
                  {ICLR} 2024, Vienna, Austria, May 7-11, 2024},
  year         = {2024},
  url          = {https://openreview.net/forum?id=jp3gWrMuIZ}
}

@article{Wang2024Understanding,
  author       = {Jiayin Wang and
                  Weizhi Ma and
                  Peijie Sun and
                  Min Zhang and
                  Jian{-}Yun Nie},
  title        = {Understanding User Experience in Large Language Model Interactions},
  journal      = {arXiv preprint},
  year         = {2024},
  url          = {https://doi.org/10.48550/arXiv.2401.08329},
  eprinttype    = {arXiv},
  eprint       = {2401.08329}
}

@article{Shaikh2023Grouding,
  author       = {Omar Shaikh and
                  Kristina Gligoric and
                  Ashna Khetan and
                  Matthias Gerstgrasser and
                  Diyi Yang and
                  Dan Jurafsky},
  title        = {Grounding or Guesswork? Large Language Models are Presumptive Grounders},
  journal      = {arXiv preprint},
  year         = {2023},
  url          = {https://doi.org/10.48550/arXiv.2311.09144},
  eprinttype    = {arXiv},
  eprint       = {2311.09144}
}

@inproceedings{Zelikman2022STaR,
  author       = {Eric Zelikman and
                  Yuhuai Wu and
                  Jesse Mu and
                  Noah D. Goodman},
  title        = {STaR: Bootstrapping Reasoning With Reasoning},
  booktitle    = {Advances in Neural Information Processing Systems 35: Annual Conference
                  on Neural Information Processing Systems 2022, NeurIPS 2022, New Orleans,
                  LA, USA, November 28 - December 9, 2022},
  year         = {2022},
  url          = {http://papers.nips.cc/paper\_files/paper/2022/hash/639a9a172c044fbb64175b5fad42e9a5-Abstract-Conference.html}
}

@article{Dong2023RAFT,
  author       = {Hanze Dong and
                  Wei Xiong and
                  Deepanshu Goyal and
                  Yihan Zhang and
                  Winnie Chow and
                  Rui Pan and
                  Shizhe Diao and
                  Jipeng Zhang and
                  Kashun Shum and
                  Tong Zhang},
  title        = {{RAFT:} Reward rAnked FineTuning for Generative Foundation Model Alignment},
  journal      = {Trans. Mach. Learn. Res.},
  year         = {2023},
  url          = {https://openreview.net/forum?id=m7p5O7zblY}
}

@inproceedings{Borsa2019Observational,
  author       = {Diana Borsa and
                  Nicolas Heess and
                  Bilal Piot and
                  Siqi Liu and
                  Leonard Hasenclever and
                  R{\'{e}}mi Munos and
                  Olivier Pietquin},
  title        = {Observational Learning by Reinforcement Learning},
  booktitle    = {Proceedings of the 18th International Conference on Autonomous Agents
                  and MultiAgent Systems, {AAMAS} '19, Montreal, QC, Canada, May 13-17,
                  2019},
  pages        = {1117--1124},
  publisher    = {International Foundation for Autonomous Agents and Multiagent Systems},
  year         = {2019},
  url          = {http://dl.acm.org/citation.cfm?id=3331811}
}

@article{Comanici2025Gemini,
  author       = {Gheorghe Comanici and
                  Eric Bieber and
                  Mike Schaekermann and
                  Ice Pasupat and
                  Noveen Sachdeva and
                  Inderjit S. Dhillon and
                  Marcel Blistein and
                  Ori Ram and
                  Dan Zhang and
                  Evan Rosen and
                  Luke Marris and
                  Sam Petulla and
                  Colin Gaffney and
                  Asaf Aharoni and
                  Nathan Lintz and
                  Tiago Cardal Pais and
                  Henrik Jacobsson and
                  Idan Szpektor and
                  Nan{-}Jiang Jiang and
                  Krishna Haridasan and
                  Ahmed Omran and
                  Nikunj Saunshi and
                  Dara Bahri and
                  Gaurav Mishra and
                  Eric Chu and
                  Toby Boyd and
                  Brad Hekman and
                  Aaron Parisi and
                  Chaoyi Zhang and
                  Kornraphop Kawintiranon and
                  Tania Bedrax{-}Weiss and
                  Oliver Wang and
                  Ya Xu and
                  Ollie Purkiss and
                  Uri Mendlovic and
                  Ila{\"{\i}} Deutel and
                  Nam Nguyen and
                  Adam Langley and
                  Flip Korn and
                  Lucia Rossazza and
                  Alexandre Ram{\'{e}} and
                  Sagar Waghmare and
                  Helen Miller and
                  Nathan Byrd and
                  Ashrith Sheshan and
                  Raia Hadsell Sangnie Bhardwaj and
                  Pawel Janus and
                  Tero Rissa and
                  Dan Horgan and
                  Sharon Silver and
                  Ayzaan Wahid and
                  Sergey Brin and
                  Yves Raimond and
                  Klemen Kloboves and
                  Cindy Wang and
                  Nitesh Bharadwaj Gundavarapu and
                  Ilia Shumailov and
                  Bo Wang and
                  Mantas Pajarskas and
                  Joe Heyward and
                  Martin Nikoltchev and
                  Maciej Kula and
                  Hao Zhou and
                  Zachary Garrett and
                  Sushant Kafle and
                  Sercan Arik and
                  Ankita Goel and
                  Mingyao Yang and
                  Jiho Park and
                  Koji Kojima and
                  Parsa Mahmoudieh and
                  Koray Kavukcuoglu and
                  Grace Chen and
                  Doug Fritz and
                  Anton Bulyenov and
                  Sudeshna Roy and
                  Dimitris Paparas and
                  Hadar Shemtov and
                  Bo{-}Juen Chen and
                  Robin Strudel and
                  David Reitter and
                  Aurko Roy and
                  Andrey Vlasov and
                  Changwan Ryu and
                  Chas Leichner and
                  Haichuan Yang and
                  Zelda Mariet and
                  Denis Vnukov and
                  Tim Sohn and
                  Amy Stuart and
                  Wei Liang and
                  Minmin Chen and
                  Praynaa Rawlani and
                  Christy Koh and
                  JD Co{-}Reyes and
                  Guangda Lai and
                  Praseem Banzal and
                  Dimitrios Vytiniotis and
                  Jieru Mei and
                  Mu Cai},
  title        = {Gemini 2.5: Pushing the Frontier with Advanced Reasoning, Multimodality,
                  Long Context, and Next Generation Agentic Capabilities},
  journal      = {arXiv preprint},
  year         = {2025},
  url          = {https://doi.org/10.48550/arXiv.2507.06261},
  eprinttype    = {arXiv},
  eprint       = {2507.06261}
}

@article{Kamath2025Gemma,
  author       = {Aishwarya Kamath and
                  Johan Ferret and
                  Shreya Pathak and
                  Nino Vieillard and
                  Ramona Merhej and
                  Sarah Perrin and
                  Tatiana Matejovicova and
                  Alexandre Ram{\'{e}} and
                  Morgane Rivi{\`{e}}re and
                  Louis Rouillard and
                  Thomas Mesnard and
                  Geoffrey Cideron and
                  Jean{-}Bastien Grill and
                  Sabela Ramos and
                  Edouard Yvinec and
                  Michelle Casbon and
                  Etienne Pot and
                  Ivo Penchev and
                  Ga{\"{e}}l Liu and
                  Francesco Visin and
                  Kathleen Kenealy and
                  Lucas Beyer and
                  Xiaohai Zhai and
                  Anton Tsitsulin and
                  R{\'{o}}bert Busa{-}Fekete and
                  Alex Feng and
                  Noveen Sachdeva and
                  Benjamin Coleman and
                  Yi Gao and
                  Basil Mustafa and
                  Iain Barr and
                  Emilio Parisotto and
                  David Tian and
                  Matan Eyal and
                  Colin Cherry and
                  Jan{-}Thorsten Peter and
                  Danila Sinopalnikov and
                  Surya Bhupatiraju and
                  Rishabh Agarwal and
                  Mehran Kazemi and
                  Dan Malkin and
                  Ravin Kumar and
                  David Vilar and
                  Idan Brusilovsky and
                  Jiaming Luo and
                  Andreas Steiner and
                  Abe Friesen and
                  Abhanshu Sharma and
                  Abheesht Sharma and
                  Adi Mayrav Gilady and
                  Adrian Goedeckemeyer and
                  Alaa Saade and
                  Alexander Kolesnikov and
                  Alexei Bendebury and
                  Alvin Abdagic and
                  Amit Vadi and
                  Andr{\'{a}}s Gy{\"{o}}rgy and
                  Andr{\'{e}} Susano Pinto and
                  Anil Das and
                  Ankur Bapna and
                  Antoine Miech and
                  Antoine Yang and
                  Antonia Paterson and
                  Ashish Shenoy and
                  Ayan Chakrabarti and
                  Bilal Piot and
                  Bo Wu and
                  Bobak Shahriari and
                  Bryce Petrini and
                  Charlie Chen and
                  Charline Le Lan and
                  Christopher A. Choquette{-}Choo and
                  CJ Carey and
                  Cormac Brick and
                  Daniel Deutsch and
                  Danielle Eisenbud and
                  Dee Cattle and
                  Derek Cheng and
                  Dimitris Paparas and
                  Divyashree Shivakumar Sreepathihalli and
                  Doug Reid and
                  Dustin Tran and
                  Dustin Zelle and
                  Eric Noland and
                  Erwin Huizenga and
                  Eugene Kharitonov and
                  Frederick Liu and
                  Gagik Amirkhanyan and
                  Glenn Cameron and
                  Hadi Hashemi and
                  Hanna Klimczak{-}Plucinska and
                  Harman Singh and
                  Harsh Mehta and
                  Harshal Tushar Lehri and
                  Hussein Hazimeh and
                  Ian Ballantyne and
                  Idan Szpektor and
                  Ivan Nardini and
                  Jean Pouget{-}Abadie and
                  Jetha Chan and
                  Joe Stanton and
                  John Wieting and
                  Jonathan Lai and
                  Jordi Orbay and
                  Joseph Fernandez and
                  Josh Newlan and
                  Ju{-}yeong Ji and
                  Jyotinder Singh and
                  Kat Black and
                  Kathy Yu and
                  Kevin Hui and
                  Kiran Vodrahalli and
                  Klaus Greff and
                  Linhai Qiu and
                  Marcella Valentine and
                  Marina Coelho and
                  Marvin Ritter and
                  Matt Hoffman and
                  Matthew Watson and
                  Mayank Chaturvedi and
                  Michael Moynihan and
                  Min Ma and
                  Nabila Babar and
                  Natasha Noy and
                  Nathan Byrd and
                  Nick Roy and
                  Nikola Momchev and
                  Nilay Chauhan and
                  Oskar Bunyan and
                  Pankil Botarda and
                  Paul Caron and
                  Paul Kishan Rubenstein and
                  Phil Culliton and
                  Philipp Schmid and
                  Pier Giuseppe Sessa and
                  Pingmei Xu and
                  Piotr Stanczyk and
                  Pouya Tafti and
                  Rakesh Shivanna and
                  Renjie Wu and
                  Renke Pan and
                  Reza Rokni and
                  Rob Willoughby and
                  Rohith Vallu and
                  Ryan Mullins and
                  Sammy Jerome and
                  Sara Smoot and
                  Sertan Girgin and
                  Shariq Iqbal and
                  Shashir Reddy and
                  Shruti Sheth and
                  Siim P{\~{o}}der and
                  Sijal Bhatnagar and
                  Sindhu Raghuram Panyam and
                  Sivan Eiger and
                  Susan Zhang and
                  Tianqi Liu and
                  Trevor Yacovone and
                  Tyler Liechty and
                  Uday Kalra and
                  Utku Evci and
                  Vedant Misra and
                  Vincent Roseberry and
                  Vlad Feinberg and
                  Vlad Kolesnikov and
                  Woohyun Han and
                  Woosuk Kwon and
                  Xi Chen and
                  Yinlam Chow and
                  Yuvein Zhu and
                  Zichuan Wei and
                  Zoltan Egyed and
                  Victor Cotruta and
                  Minh Giang and
                  Phoebe Kirk and
                  Anand Rao and
                  Jessica Lo and
                  Erica Moreira and
                  Luiz Gustavo Martins and
                  Omar Sanseviero and
                  Lucas Gonzalez and
                  Zach Gleicher and
                  Tris Warkentin and
                  Vahab Mirrokni and
                  Evan Senter and
                  Eli Collins and
                  Joelle K. Barral and
                  Zoubin Ghahramani and
                  Raia Hadsell and
                  Yossi Matias and
                  D. Sculley and
                  Slav Petrov and
                  Noah Fiedel and
                  Noam Shazeer and
                  Oriol Vinyals and
                  Jeff Dean and
                  Demis Hassabis and
                  Koray Kavukcuoglu and
                  Cl{\'{e}}ment Farabet and
                  Elena Buchatskaya and
                  Jean{-}Baptiste Alayrac and
                  Rohan Anil and
                  Dmitry (Dima) Lepikhin and
                  Sebastian Borgeaud and
                  Olivier Bachem and
                  Armand Joulin and
                  Alek Andreev and
                  Cassidy Hardin and
                  Robert Dadashi and
                  L{\'{e}}onard Hussenot},
  title        = {Gemma 3 Technical Report},
  journal      = {arXiv preprint},
  year         = {2025},
  url          = {https://doi.org/10.48550/arXiv.2503.19786},
  eprinttype    = {arXiv},
  eprint       = {2503.19786}
}

@inproceedings{Krishnamurthy2024Can,
  author       = {Akshay Krishnamurthy and
                  Keegan Harris and
                  Dylan J. Foster and
                  Cyril Zhang and
                  Aleksandrs Slivkins},
  title        = {Can large language models explore in-context?},
  booktitle    = {Advances in Neural Information Processing Systems 38: Annual Conference
                  on Neural Information Processing Systems 2024, NeurIPS 2024, Vancouver,
                  BC, Canada, December 10 - 15, 2024},
  year         = {2024},
  url          = {http://papers.nips.cc/paper\_files/paper/2024/hash/d951f73c521d069fefbb73396df01424-Abstract-Conference.html}
}

@inproceedings{Dai2024In-context,
  author       = {Zhenwen Dai and
                  Federico Tomasi and
                  Sina Ghiassian},
  title        = {In-context Exploration-Exploitation for Reinforcement Learning},
  booktitle    = {The Twelfth International Conference on Learning Representations,
                  {ICLR} 2024, Vienna, Austria, May 7-11, 2024},
  year         = {2024},
  url          = {https://openreview.net/forum?id=uIKZSStON3}
}

@article{White2023A,
  author       = {Jules White and
                  Quchen Fu and
                  Sam Hays and
                  Michael Sandborn and
                  Carlos Olea and
                  Henry Gilbert and
                  Ashraf Elnashar and
                  Jesse Spencer{-}Smith and
                  Douglas C. Schmidt},
  title        = {A Prompt Pattern Catalog to Enhance Prompt Engineering with ChatGPT},
  journal      = {arXiv preprint},
  year         = {2023},
  url          = {https://doi.org/10.48550/arXiv.2302.11382},
  eprinttype    = {arXiv},
  eprint       = {2302.11382}
}

@article{Chu2025SFT,
  author       = {Tianzhe Chu and
                  Yuexiang Zhai and
                  Jihan Yang and
                  Shengbang Tong and
                  Saining Xie and
                  Dale Schuurmans and
                  Quoc V. Le and
                  Sergey Levine and
                  Yi Ma},
  title        = {{SFT} Memorizes, {RL} Generalizes: {A} Comparative Study of Foundation
                  Model Post-training},
  journal      = {arXiv preprint},
  year         = {2025},
  url          = {https://doi.org/10.48550/arXiv.2501.17161},
  eprinttype    = {arXiv},
  eprint       = {2501.17161}
}

@inproceedings{Ostrovski2021The,
  author       = {Georg Ostrovski and
                  Pablo Samuel Castro and
                  Will Dabney},
  title        = {The Difficulty of Passive Learning in Deep Reinforcement Learning},
  booktitle    = {Advances in Neural Information Processing Systems 34: Annual Conference
                  on Neural Information Processing Systems 2021, NeurIPS 2021, December
                  6-14, 2021, virtual},
  pages        = {23283--23295},
  year         = {2021},
  url          = {https://proceedings.neurips.cc/paper/2021/hash/c3e0c62ee91db8dc7382bde7419bb573-Abstract.html}
}

@inproceedings{Kirk2024Understanding,
  author       = {Robert Kirk and
                  Ishita Mediratta and
                  Christoforos Nalmpantis and
                  Jelena Luketina and
                  Eric Hambro and
                  Edward Grefenstette and
                  Roberta Raileanu},
  title        = {Understanding the Effects of {RLHF} on {LLM} Generalisation and Diversity},
  booktitle    = {The Twelfth International Conference on Learning Representations,
                  {ICLR} 2024, Vienna, Austria, May 7-11, 2024},
  year         = {2024},
  url          = {https://openreview.net/forum?id=PXD3FAVHJT}
}

@article{Cook2025Programming,
  author       = {Jonathan Cook and
                  Silvia Sapora and
                  Arash Ahmadian and
                  Akbir Khan and
                  Tim Rockt{\"{a}}schel and
                  Jakob N. Foerster and
                  Laura Ruis},
  title        = {Programming by Backprop: LLMs Acquire Reusable Algorithmic Abstractions
                  During Code Training},
  journal      = {arXiv preprint},
  year         = {2025},
  url          = {https://doi.org/10.48550/arXiv.2506.18777},
  eprinttype    = {arXiv},
  eprint       = {2506.18777}
}

@article{Vygotski1929The,
  author       = {Lev S. Vygotski},
  title        = {The Problem of the Cultural Development of the Child},
  journal      = {The Pedagogical Seminary and Journal of Genetic Psychology},
  publisher    = {Taylor \& Francis Group},
  pages        = {415--434},
  volume       = {6},
  year         = {1929},
  url          = {https://www.tandfonline.com/doi/pdf/10.1080/08856559.1929.10532201}
}

@article{Ahmad2025OpenCodeInstruct,
  author       = {Wasi Uddin Ahmad and
                  Aleksander Ficek and
                  Mehrzad Samadi and
                  Jocelyn Huang and
                  Vahid Noroozi and
                  Somshubra Majumdar and
                  Boris Ginsburg},
  title        = {OpenCodeInstruct: {A} Large-scale Instruction Tuning Dataset for Code
                  LLMs},
  journal      = {arXiv preprint},
  year         = {2025},
  url          = {https://doi.org/10.48550/arXiv.2504.04030},
  eprinttype    = {arXiv},
  eprint       = {2504.04030}
}

@article{Yang2025Qwen3,
  author       = {An Yang and
                  Anfeng Li and
                  Baosong Yang and
                  Beichen Zhang and
                  Binyuan Hui and
                  Bo Zheng and
                  Bowen Yu and
                  Chang Gao and
                  Chengen Huang and
                  Chenxu Lv and
                  Chujie Zheng and
                  Dayiheng Liu and
                  Fan Zhou and
                  Fei Huang and
                  Feng Hu and
                  Hao Ge and
                  Haoran Wei and
                  Huan Lin and
                  Jialong Tang and
                  Jian Yang and
                  Jianhong Tu and
                  Jianwei Zhang and
                  Jian Yang and
                  Jiaxi Yang and
                  Jingren Zhou and
                  Junyang Lin and
                  Kai Dang and
                  Keqin Bao and
                  Kexin Yang and
                  Le Yu and
                  Lianghao Deng and
                  Mei Li and
                  Mingfeng Xue and
                  Mingze Li and
                  Pei Zhang and
                  Peng Wang and
                  Qin Zhu and
                  Rui Men and
                  Ruize Gao and
                  Shixuan Liu and
                  Shuang Luo and
                  Tianhao Li and
                  Tianyi Tang and
                  Wenbiao Yin and
                  Xingzhang Ren and
                  Xinyu Wang and
                  Xinyu Zhang and
                  Xuancheng Ren and
                  Yang Fan and
                  Yang Su and
                  Yichang Zhang and
                  Yinger Zhang and
                  Yu Wan and
                  Yuqiong Liu and
                  Zekun Wang and
                  Zeyu Cui and
                  Zhenru Zhang and
                  Zhipeng Zhou and
                  Zihan Qiu},
  title        = {Qwen3 Technical Report},
  journal      = {arXiv preprint},
  volume       = {abs/2505.09388},
  year         = {2025},
  url          = {https://doi.org/10.48550/arXiv.2505.09388},
  eprinttype    = {arXiv},
  eprint       = {2505.09388}
}

\appendix

\newpage
\section{Qwen Results}
\label{ap:qwen}

\begin{figure}[h!]
\begin{center}
\includegraphics[width=7cm, height=5cm]{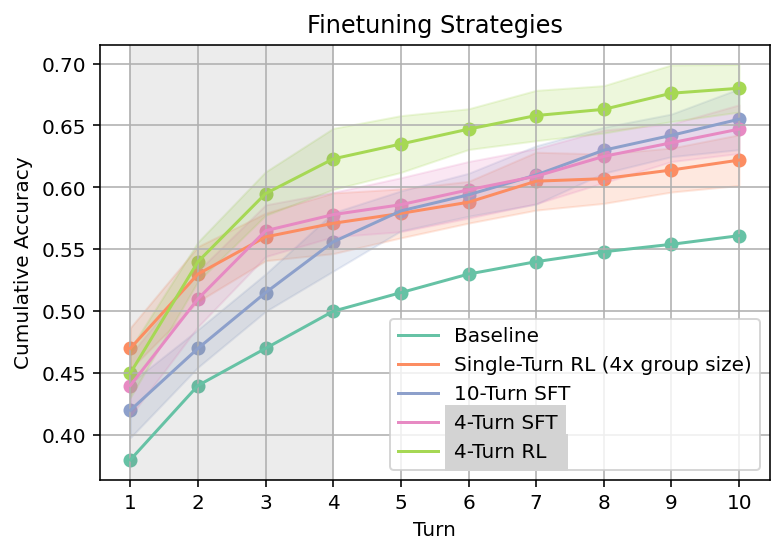}
\end{center}
\vspace{-0.3cm}
\caption{Comparing different multi-turn and single-turn finetuning strategies on Omni-MATH.}
\label{fig:qwen_id}
\end{figure}

\begin{figure}[h!]
\begin{center}
\includegraphics[width=12cm, height=5cm]{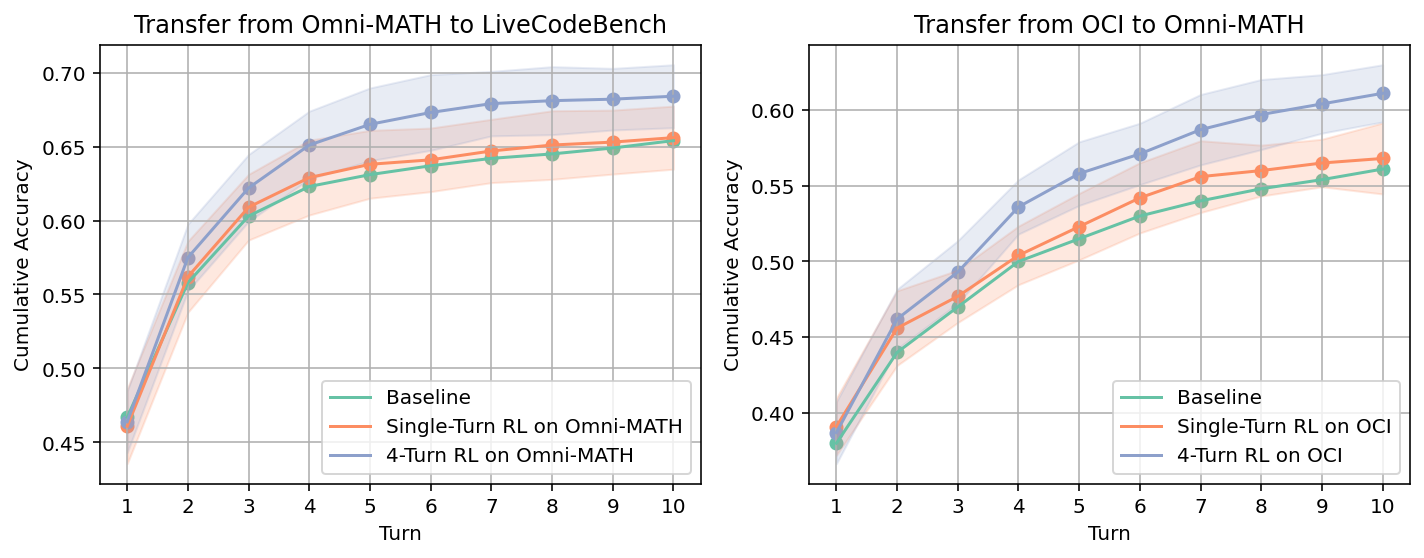}
\end{center}
\vspace{-0.3cm}
\caption{Evaluating transfer of the ability to learn from language feedback between math and code domains. Left: Training on Omni-MATH and evaluating on LiveCodeBench; Right: Training on OpenCodeInstruct and evaluating on Omni-MATH.}
\label{fig:qwen_transfer}
\end{figure}

\end{document}